# Differentiating Between Human-Written and AI-Generated Texts Using Automatically Extracted Linguistic Features

Georgios P. Georgiou[1,2]

[1]Department of Languages and Literature, University of Nicosia, Nicosia 2417, Cyprus

[2]Phonetic Lab, University of Nicosia, Nicosia 2417, Cyprus

**Abstract**

While extensive research has focused on ChatGPT in recent years, very few studies have systematically quantified and compared linguistic features between human-written and artificial intelligence (AI)-generated language. This exploratory study aims to investigate how various linguistic components are represented in both types of texts, assessing AI's ability to emulate human writing. Using human-authored essays as a benchmark, we prompted ChatGPT to generate essays of equivalent length. These texts were analyzed using Open Brain AI, an online computational tool, to extract measures of phonological, morphological, syntactic, and lexical constituents. Despite AI-generated texts appearing to mimic human speech, the results revealed significant differences across multiple linguistic features such as specific types of consonants, nouns, adjectives, pronouns, adjectival/prepositional modifiers, and use of difficult words, among others. These findings underscore the importance of integrating automated tools for efficient language assessment, reducing time and effort in data analysis. Moreover, they emphasize the necessity for enhanced training methodologies to improve AI's engineering capacity for producing more human-like text.

Keywords:

ChatGPT; essays; Open Brain AI; linguistic features

**Graphical Abstract**

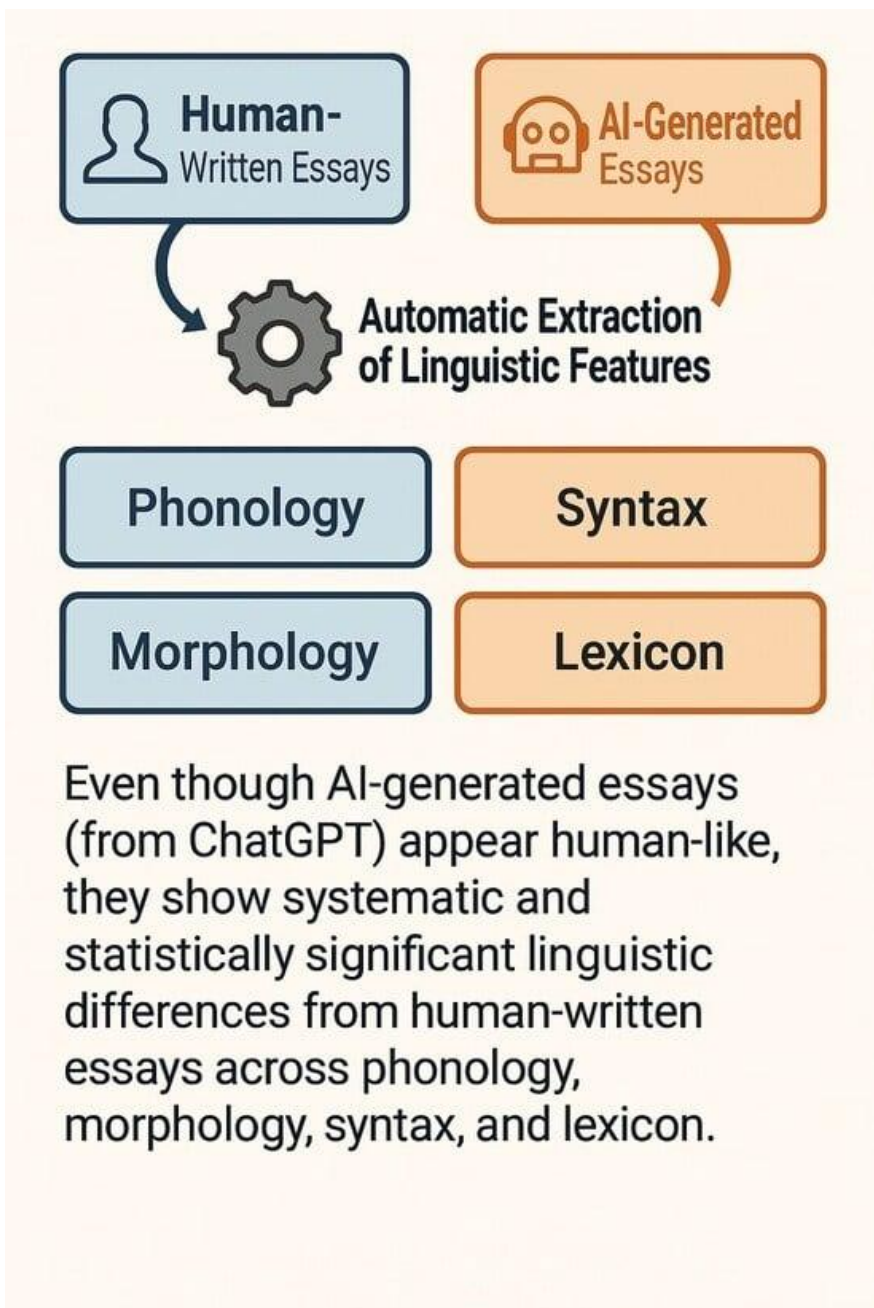

## 1. Introduction

The revolutionary advancements in artificial intelligence (AI) and natural language processing (NLP) have given rise to increasingly sophisticated and capable language models known as large language models (LLMs) [1]. These LLMs, a subset of generative AI, generate novel text by leveraging patterns learned from large corpora, enabling machines to comprehend and produce human language [2]. ChatGPT is a conversational interface built by OpenAI on top of Generative Pre-trained Transformer (GPT) models (e.g., GPT-3.5, GPT-4, and GPT-5), which produce contextually appropriate responses to user prompts and support applications across education, healthcare, language learning, and customer service [3,4,5,6].

One important advantage of ChatGPT is the incorporation of human feedback during training—commonly via reinforcement learning from human feedback —which helps the system better align responses with user intent [7]. Early deployments of ChatGPT were powered by GPT-3 lineage, which was reported to have ~175 billion parameters and to have been trained for a total of ~300 billion tokens [8]. By learning the subtleties of human language from these large-scale datasets, ChatGPT can produce text that closely mimics human writing [9].

However, AI language does not follow the exact patterns found in human language. A small body of work compared human and AI languages to detect potential differences. For example, a recent qualitative study conducted in Cyprus by Alexander et al. [10] explored how ESL lecturers identify AI-generated academic texts and the challenges they face in doing so. Six lecturers evaluated four C1-level essays, with their judgments compared to results from AI detection tools. The study found that lecturers often relied on a “deficit model” of assessment, interpreting linguistic accuracy and sophistication as indicators of AI authorship. The authors concluded that ChatGPT-generated essays were highly recognizable compared to student-written essays since they presented divergent language patterns. Herbold et al. [11] examined human-written versus ChatGPT-generated essays in an attempt to identify the linguistic devices that are characteristic of student versus AI-generated content, among others. The AI-generated essays were assessed by teachers. The results indicated significant linguistic distinctions between human-written and AI-generated content. AI-generated essays exhibited a high degree of structural uniformity, exemplified by identical introductions to concluding sections across all ChatGPT essays. Furthermore, initial sentences in each essay tended to start with a generalized statement using key concepts from the essay topics, reflecting a structured approach typical of argumentative essays. This contrasts with human-written essays, which display greater variability in adhering to such structural guidelines on the linguistic surface. In another study, Cai et al. [12] examined how LLMs, including ChatGPT and Vicuna, effectively replicate human language. Various prompts examining phonetic, syntactic, semantic, and discourse patterns were given directly to the chatbots. The models were found to replicate human language well across all levels. However, some discrepancies occurred since, unlike humans, neither model showed a preference for using shorter words to express less informative content, nor did it utilize context to resolve syntactic ambiguities. A more comprehensive comparative examination of the human and AI languages, with a special focus on particular linguistic features, is needed to better understand the differences between them.

Recent comparisons of human and LLM writing converge on interpretable stylistic footprints that make model text detectable even when it is fluent. Beyond domain-specific studies in education and medicine (e.g., [11,13]), three families of indicators recur. The first is part-of-speech (POS) balance. AI outputs often show nominal loading—more nouns with comparatively fewer pronouns and auxiliaries—yielding denser, more templatic clauses; human texts exhibit richer functional morphology supporting reference, perspective, and tense/aspect variation (cf. [10,11,13]). Probability-based detectors such as DetectGPT exploit distributional regularities by testing whether candidate text lies in regions of characteristic probability curvature under a model [14]; Fast-DetectGPT accelerates this approach while preserving zero-shot detection performance [15]. The second indicator is about lexical density and vocabulary profile. Model text tends toward higher lexical density and more "difficult" content words, especially in academic registers, whereas human writers deploy more function words and audience-tuned paraphrase (cf. [10,16,17]). Recent stylometric work confirms density and vocabulary shifts across LLMs in expository and creative settings [18,19]. The third indicator is cohesion and discourse scaffolding. Humans rely more on cohesive devices—pronouns, prepositional and adverbial modifiers, and varied connectives—to maintain referential chains and signal discourse relations; AI text often shows overt conjunctive framing but weaker referential maintenance (cf. [11,16,17,20,21]). At the same time, robustness studies caution that detectors are fragile to paraphrase and adversarial rewriting, and that watermark-avoidance can erode reliability—limits that argue for transparent, interpretable features rather than sole reliance on black-box signals [22].

### *1.1. Automatic Elicitation of Linguistic Features*

Analyzing linguistic features in language, speech, and communication provides valuable insights into linguistic choices and aids in language assessment. Such analyses have heavily relied on manual assessments in both typical and clinical populations. For example, the mapping of speech production divergences of second language speakers often requires the collection of speech recordings, the segmentation of sounds, and the implementation of statistical analysis (see [23]). In clinical settings involving children with developmental language disorder, extracting speech patterns usually demands the use of a narrative assessment tool followed by audio analysis (see [24]). These methods, although reliable, can be unwieldy and time-consuming, potentially causing stress for patients or students undergoing the assessments [25].

The evolution of AI technologies has brought automated computational applications to the forefront, simplifying the extraction of speech and language measures. These tools utilize machine learning technologies, such as deep neural networks and NLP, to provide algorithms for linguistic analysis and pattern interpretation. Open Brain AI [26] is an open-source online computational platform application designed to provide automated linguistic and cognitive assessments. It serves researchers, clinicians, and educators by streamlining their daily tasks through advanced AI methods and tools. Educators can utilize the application to evaluate students' speech and language, extract meaningful markers from essays and other materials, estimate performance, and assess the effectiveness of teaching methodologies. For clinicians, Open Brain AI automates the analysis of spoken and written language, offering valuable linguistic insights into the language of patients. Since language can be indicative of potential speech, language, and communication disorders, early screening and assessment can be decisive for diagnosis and treatment [27]. Researchers may benefit from Open Brain AI by generating quantitative measures of speech, language, and communication that can assist their research.

The analysis provided by the application involves the objective measurement of written language production features, allowing the comparison of an individual with a targeted population across various linguistic domains. It specifically analyzes texts or transcripts generated from the speech-to-text module, conducting assessments in the following linguistic domains:

- Phonology: Measures include the number and ratios of syllables, vowels, words with primary and secondary stress, consonants per place and manner of articulation, and voiced and voiceless consonants.
- Morphology: Includes the counts and ratios of parts of speech (e.g., verbs, nouns, adjectives, adverbs, conjunctions, etc.) relative to the total number of words.
- Syntax: Calculates the counts and ratios of syntactic constituents (e.g., modifiers, case markers, direct objects, nominal subjects, predicates, etc.).
- Lexicon: Provides metrics such as the total number of words, hapax legomena (words that occur once), Type Token Ratio (TTR), and others.
- Semantics: Estimates counts and ratios of semantic entities within the text (e.g., persons, dates, locations, etc.).
- Readability Measures: Assessments of text readability and grammatical structure.

Our work is grounded in stylometry and computational linguistics, traditions that use measurable linguistic features, spanning syntax, morphology, lexis, and function-word profiles, to distinguish authorial style and attribute authorship with interpretable cues rather than opaque signals [28,29]. We extend prior POS/lexical-density/cohesion work in two ways. First, we add phonology-adjacent indices derived from orthography (place/manner, voicing, primary stress), dimensions largely absent from human–AI stylometry and probability-curvature detection. Second, we use nonparametric hypothesis testing with regularized classification combined with robust rank-based tests and sparse penalized regression, which yield signed, sparse coefficients per level, complementing curvature-based detectors and isolating robust cues. Practically, our task-matched, length-controlled essay design and single-tool workflow provide a reproducible, interpretable pipeline.

### *1.2. This Study*

This study aims to investigate the representation of various linguistic constituents in both human-written and AI-generated texts. The main objective is to find potential differences between human and AI essay texts in the occurrence of various phonological, morphological, syntactic, and lexical components. This is one of the very few studies to focus on how particular features can discern human-written and AI-generated texts. The present study employs a dataset of ten texts—five human-written and five generated by ChatGPT—as a deliberate design choice to facilitate a controlled, feature-rich comparison between human and AI-generated language, with the analysis of 9201 linguistic observations. Such controlled conditions are particularly valuable in exploratory studies aiming to identify systematic patterns in linguistic behavior rather than to make broad population-level generalizations. Small, well-curated datasets have a long-standing precedent in linguistic and computational research, particularly when the focus is on fine-grained feature analysis or proof-of-concept investigations. This study follows that tradition by offering detailed insights into key linguistic differences, while laying the groundwork for subsequent large-scale analyses. The findings are not intended to be exhaustive, but rather to serve as an empirically grounded starting point for further inquiry. A more extensive and diverse dataset would certainly strengthen the generalizability of future work, and we highlight this as a productive direction for follow-up research.

Such an investigation is crucial for advancing multiple aspects of language technology and its applications. It is essential for understanding AI capabilities and limitations, enabling the refinement of algorithms to produce more natural and coherent language. By identifying where AI text diverges from human norms, researchers can improve training methods and design better models, enhancing the quality of AI-generated content. These improvements have broad applications in NLP tasks such as machine translation, text summarization, and dialogue systems, and can significantly benefit content creation in fields such as marketing, journalism, education, and health services. Furthermore, understanding linguistic discrepancies enhances user experience and fosters trust and acceptance of AI technologies by ensuring that generated content meets human expectations. Moreover, this study will effectively highlight the role of LLMs in supporting the development of automated linguistic assessments through a user-friendly tool.

The remainder of this paper is organized as follows. Section 2 describes the methodological framework, including data selection, preprocessing, and statistical procedures. Section 3 presents the results of the linguistic comparison between human-written and AI-generated texts across phonological, morphological, syntactic, and lexical domains. Section 4 discusses these findings in relation to existing literature, highlighting their interpretability and implications. Finally, Section 5 summarizes the main conclusions, acknowledges limitations, and outlines directions for future research.

## 2. Methodology

### *2.1. Procedure*

Ten essays were included in the analysis. Five were authentic IELTS writing task samples produced by professional English teachers, ensuring grammatical accuracy and consistency with B2–C1 proficiency standards. These texts covered diverse yet comparable topics—education, technology, society, environment, and the arts—reflecting general argumentative writing genres. The mean text length was 310 words (*SD* = 48.19), with an average of 17.2 sentences per essay (*SD* = 3.77).

To create a matched set, we used ChatGPT (GPT-4, March 2024 version) to generate five additional texts based on the same IELTS task prompts. The model was accessed through the ChatGPT web interface, which does not allow manual adjustment of decoding parameters such as temperature, top-p, presence penalty, or frequency penalty. Therefore, text generation relied on the platform's default internal settings. The target length was set based on the length of each human text to parallel human-authored texts. Prompts explicitly included task instructions (e.g., "Write an IELTS Task 2 essay on [topic], approximately [number] words, formal academic register") and an in-context directive specifying register and structure ("Use academic English and structured argumentation typical of IELTS Band 8 responses"). No external context or examples were provided to maintain independent generation for each essay.

All texts underwent light preprocessing before analysis. We applied light, content-preserving preprocessing: Unicode normalization with UTF-8 encoding; whitespace normalization (trimming leading/trailing spaces, collapsing multiple spaces, and standardizing line breaks to \n); standardization of quotes and dashes (smart → straight quotes; em/en dashes → hyphen) without deleting punctuation; and removal of non-text artifacts (e.g., HTML tags, copy/paste artifacts) while preserving words, punctuation, and case. No additional normalization was applied to preserve the original linguistic structure.

Text samples were then uploaded to the Open Brain AI beta web platform, using the linguistic profiling module with subcomponents for readability, phonology, morphology, syntax, and lexis. The platform applied a built-in NLP pipeline (tokenization, lemmatization, POS tagging, parsing) to compute morphological, syntactic, lexical, phonology-adjacent, and readability measures. Outputs were provided as a per-document feature table, which we analyzed without modifying the platform's default settings.

The automatic extraction produced 94 linguistic measures: 22 phonological, 15 morphological, 44 syntactic, and 13 lexical. In total, 9201 linguistic observations were recorded across domains, providing sufficient granularity for inferential comparison. For subsequent analysis, we included only features corresponding to counts or normalized frequencies of linguistic constituents (e.g., word, syllable, clause, or phrase units) and those exhibiting substantial presence across at least 80% of texts. This criterion ensured the inclusion of robust, representative variables rather than sparse or idiosyncratic measures. Readability indices were analyzed descriptively.

Although the sample size was limited, the controlled pairing of human and AI-generated texts under matched task conditions allowed for a fine-grained, feature-level comparison of linguistic patterns.

### *2.2. Statistical Analysis*

All statistical analyses were performed in R [30]. For each linguistic feature, we compared the ratios of AI-generated and human-produced texts using the Wilcoxon rank-sum test, a nonparametric test that does not assume normality and is robust for small sample sizes. The test statistic (W) and corresponding *p*-value were extracted for each variable. To quantify the magnitude and direction of differences, we computed Hedges' g (an unbiased estimate of Cohen's *d*) with 95% confidence intervals, using the effsize package. Confidence intervals were derived via the rstatix package. This approach provides both inferential and effect-size information for each linguistic variable across four linguistic domains: phonology, morphology, syntax, and lexicon. When appropriate, *p*-values were corrected for multiple comparisons using the Holm method to control the family-wise error rate.

For each linguistic domain—phonology, morphology, syntax, and lexicon—a separate Least Absolute Shrinkage and Selection Operator (LASSO) regression was fitted to identify which features best discriminated AI-generated from human-produced texts. Within each model, the binary outcome variable was source (AI/Human), and predictors were all features belonging to that level. All variables were analyzed on their original scales. Each LASSO model was estimated with the glmnet package in R (family = "binomial", $\alpha = 1$), using internal cross-validation to select the penalty parameter that minimized classification error ($\lambda_{min}$). Coefficients with non-zero weights at this value of $\lambda$ were interpreted as features contributing uniquely to class discrimination. Positive weights indicate features more typical of AI texts, whereas negative weights indicate features more typical of human texts. Figure 1 illustrates an overview of the study workflow and analysis pipeline.

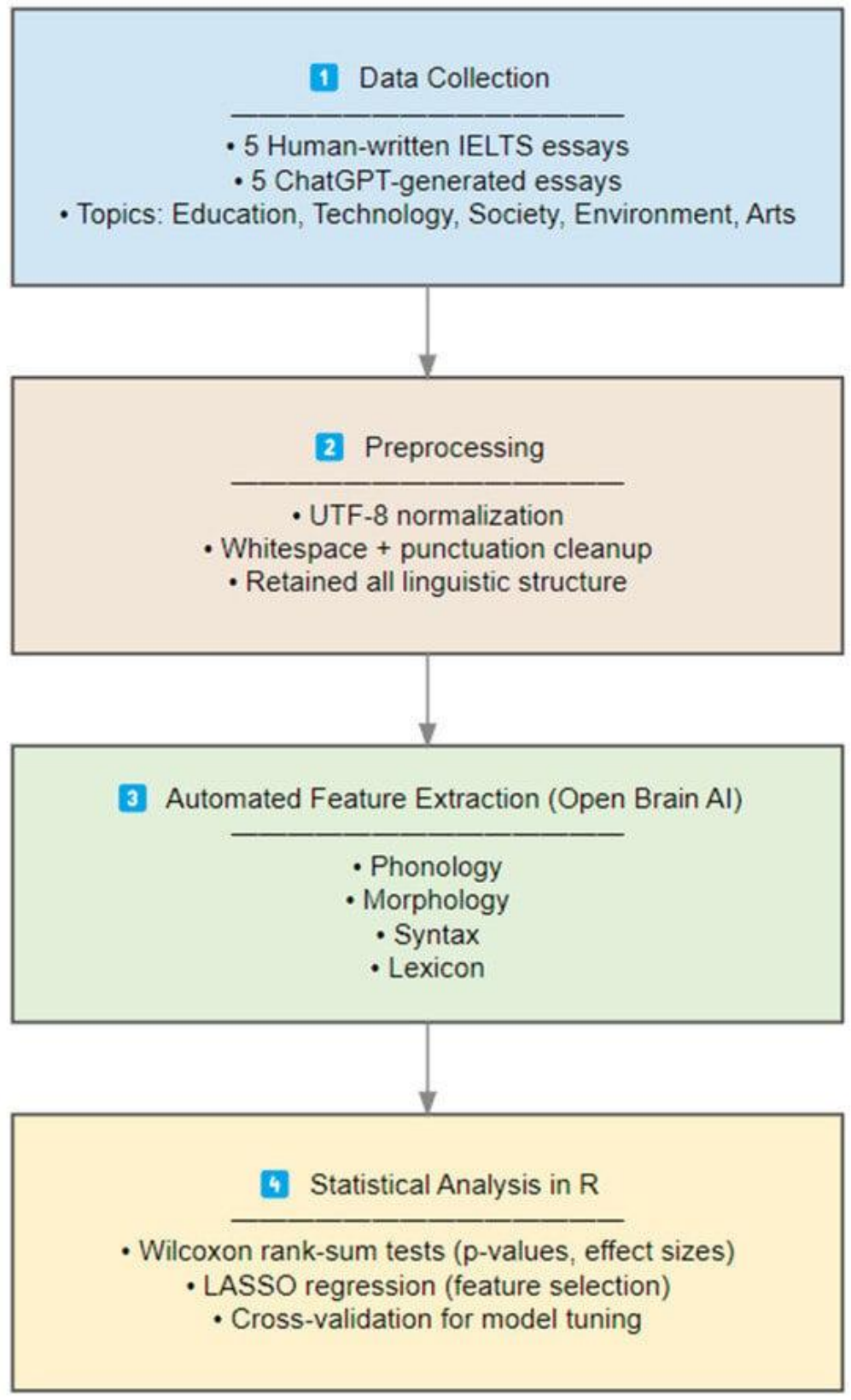

**Figure 1.** Overview of the study workflow and analysis pipeline.

## 3. Results

Table 1 presents the average readability measures across the human and AI texts. To investigate potential differences between human and AI texts, we used statistical analyses. The Wilcoxon tests demonstrated significant differences for nasals, alveolars, dentals, labiodentals, and voiceless consonants between human and AI texts, indicating a tendency of AI to favor the first two categories and voiceless consonants. The effect sizes were large. With respect to morphology, there were significant differences in the ratios of adjectives, adpositions, auxiliaries, coordinating conjunctions, nouns, and pronouns, all with large effect sizes. More specifically, the AI text used more coordinating conjunctions and nouns. Regarding syntax, significant differences and large effect sizes were found for adjectival modifiers, conjuncts, object prepositions, and prepositional modifiers between human and AI texts, with AI producing proportionally more tokens for adjectival modifiers and conjuncts. Lastly, the tests exhibited significant differences in the use of difficult words, content words, and function words between the two types of texts; these effects were large. The AI text used a greater number of difficult and content words, while the human text used a greater number of function words. Table 2 shows the results of the Wilcoxon tests. Figure 2, Figure 3, Figure 4 and Figure 5 illustrate the ratios and standard deviations of phonological, morphological, syntactic, and lexical components in both the human-written and the AI-generated texts.

Table 1. Average readability measures across the human and AI texts

| measure | human | AI |
| --- | --- | --- |
| Estimated Reading Time (sec) | 22.462 | 26.764 |
| Flesch Reading Ease | 55.352 | 28.444 |
| Flesch-Kincaid Grade Level | 10.366 | 13.795 |
| Gunning Fog Index | 12.537 | 16.391 |
| Coleman-Liau Index | 11.651 | 17.306 |
| Automated Readability Index | 11.498 | 15.463 |
| Smog Index | 12.482 | 15.690 |
| Linsear Write Formula | 2.645 | 2.715 |
| Passive Sentences Percent | 33.593 | 11.055 |
| Dale Chall Readability Score | 9.498 | 12.176 |
| Difficult words | 95.800 | 146.00 |

**Table 2.** Results of the Wilcoxon rank-sum test

| **level** | **feature** | ***p*-value** | **95% CI** | **W** | **Hedges' *g*** |
|---|---|---|---|---|---|
| phonology | approximant | 1.0000 | –1.48 – 1.16 | 12 | -0.16 |
| | fricative | 0.4034 | –1.95 – 0.75 | 8 | –0.60 |
| | lateral | 1.0000 | –1.12 – 1.53 | 13 | 0.20 |
| | nasal | 0.0216 | 0.01 – 2.99 | 24 | 1.50 |
| | plosive | 0.4034 | –0.60 – 2.13 | 17 | 0.77 |
| | alveolar | 0.0122 | 0.80 – 4.37 | 25 | 2.58 |
| | dental | 0.0216 | –3.83 – –0.51 | 1 | –2.17 |
| | labiodental | 0.0216 | –3.54 – –0.34 | 1 | –1.94 |
| | postalveolar | 0.2963 | –0.86 – 1.81 | 18 | 0.47 |
| | voiced | 0.4034 | –1.83 – 0.84 | 8 | –0.50 |
| | voiceless | 0.0367 | 0.00 – 2.98 | 23 | 1.49 |
| | primary stress | 0.2963 | –2.40 – 0.40 | 7 | –1.00 |
| | secondary stress | 0.8345 | –1.35 – 1.20 | 11 | –0.03 |
| morphology | adjective | 0.0367 | 0.25 – 3.38 | 23 | 1.81 |
| | adposition | 0.0212 | –3.62 – –0.39 | 1 | –2.00 |
| | adverb | 0.6752 | –1.15 – 1.49 | 15 | 0.17 |
| | auxiliary | 0.0122 | –3.97 – –0.59 | 0 | –2.28 |
| | coordinating conjunction | 0.0356 | 0.03 – 3.02 | 23 | 1.52 |
| | noun | 0.0122 | 0.78 – 4.33 | 25 | 2.56 |
| | particle | 0.7526 | –1.45 – 1.18 | 10.5 | –0.14 |
| | pronoun | 0.0119 | –5.20 – –1.22 | 0 | –3.21 |
| | subordinating conjunction | 0.2059 | –2.21 – 0.54 | 6 | –0.84 |
| | verb | 0.8345 | –1.23 – 1.40 | 14 | 0.09 |
| syntax | adjectival modifier | 0.0278 | 0.15 – 3.22 | 23.5 | 1.68 |
| | adverbial modifier | 1.0000 | –1.39 – 1.25 | 12 | –0.07 |
| | conjunct | 0.0160 | 0.05 – 3.05 | 24.5 | 1.55 |
| | determiner | 0.1732 | –2.23 – 0.52 | 5.5 | –0.85 |
| | direct object | 0.0662 | –0.24 – 2.62 | 21.5 | 1.19 |
| | nominal subject | 0.2101 | –0.39 – 2.41 | 19 | 1.01 |
| | object of preposition | 0.0465 | –2.93 – 0.03 | 2.5 | –1.45 |
| | prepositional modifier | 0.0367 | –-3.22 – 0.16 | 2 | –1.69 |
| | root | 0.1732 | –0.69 – 2.02 | 19.5 | 0.66 |
| lexicon | difficult word | 0.0122 | 1.02 – 4.79 | 15 | 2.90 |
| | content word | 0.0122 | 0.88 – 4.51 | 25 | 2.69 |
| | function word | 0.0122 | –4.78 – –1.01 | 0 | –2.90 |

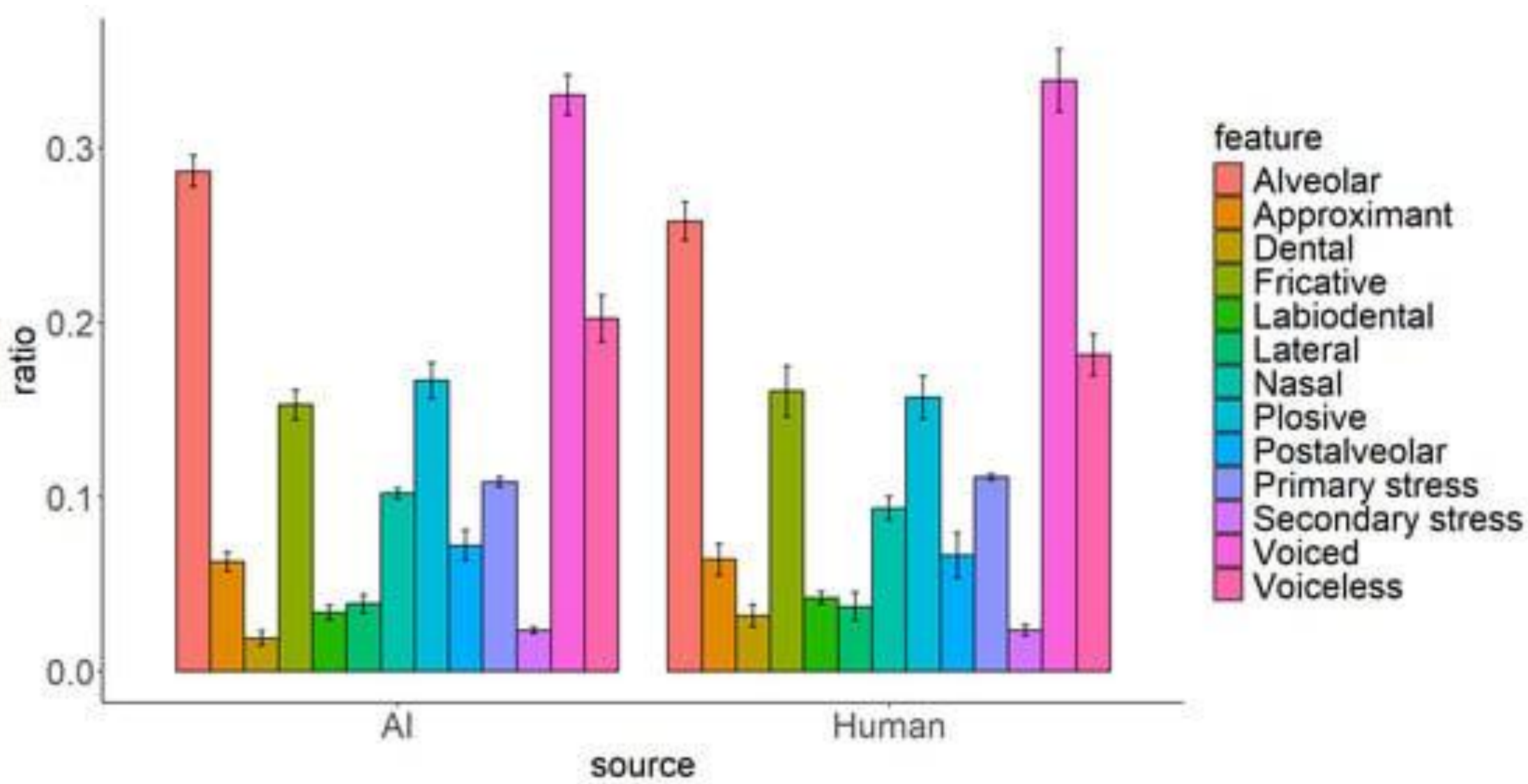

**Figure 2.** Ratios of phonological components in both the human-written and the AI-generated texts.

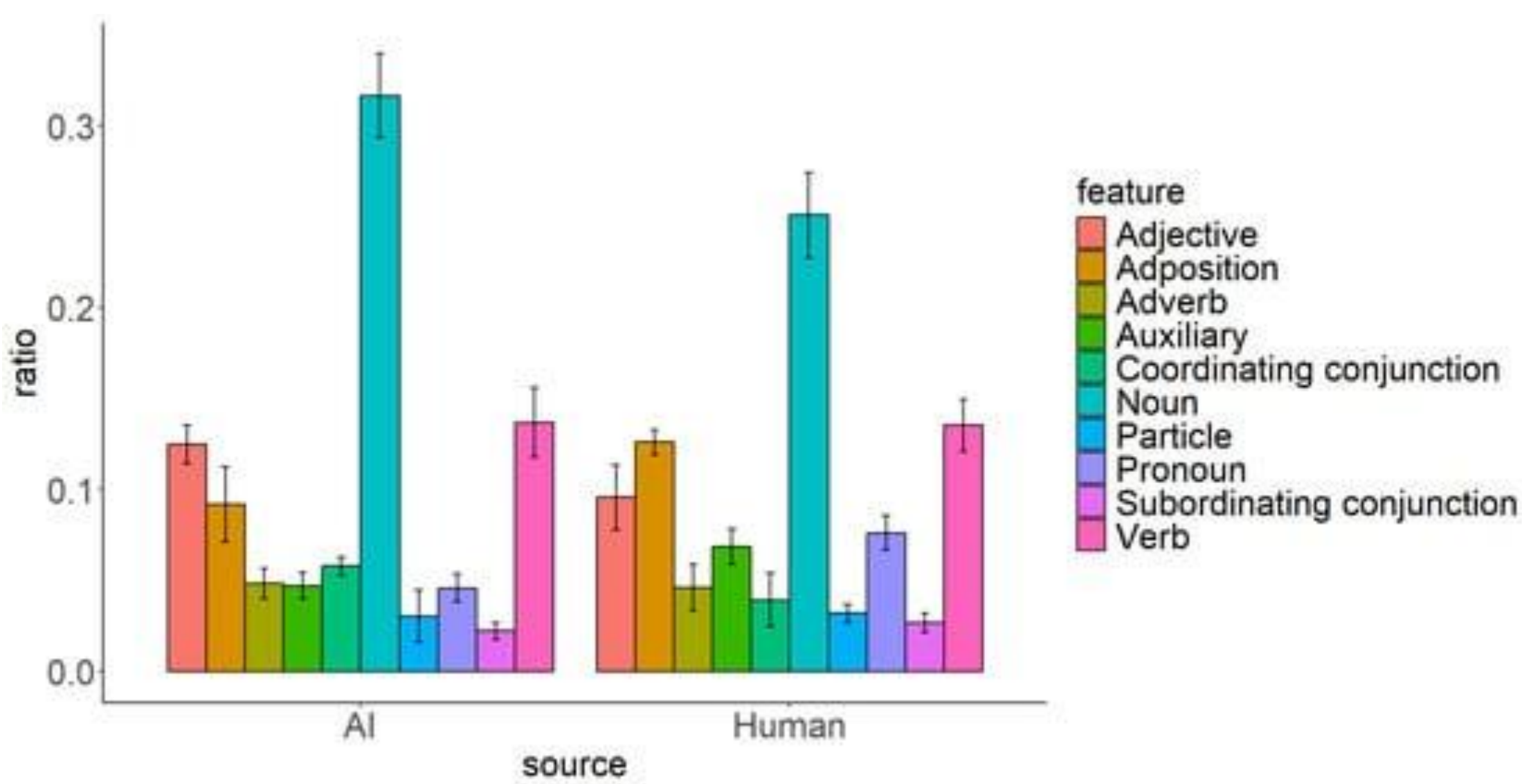

**Figure 3.** Ratios of morphological components in both the human-written and the AI-generated texts.

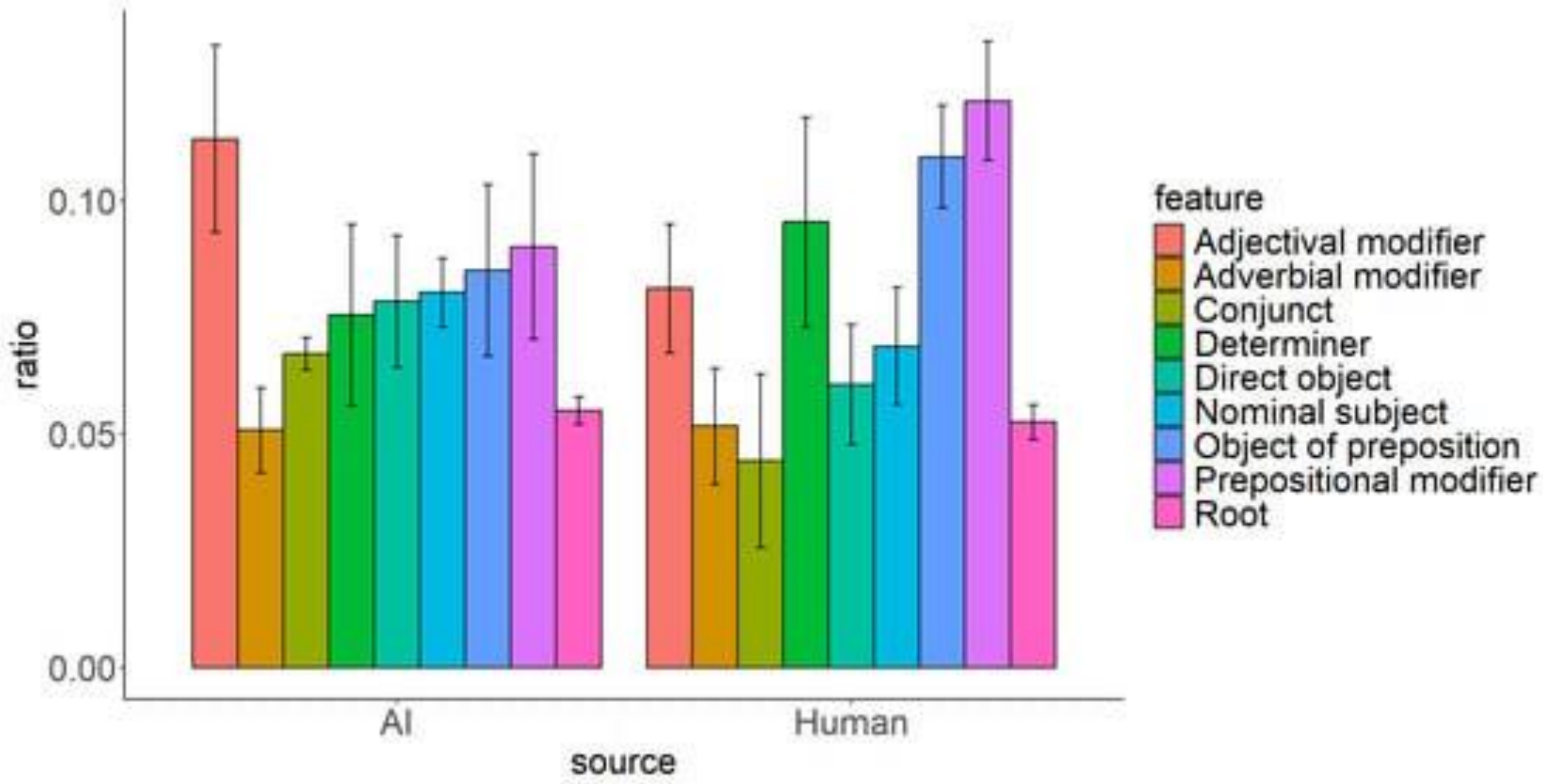

**Figure 4.** Ratios of syntactic components in both the human-written and the AI-generated texts.

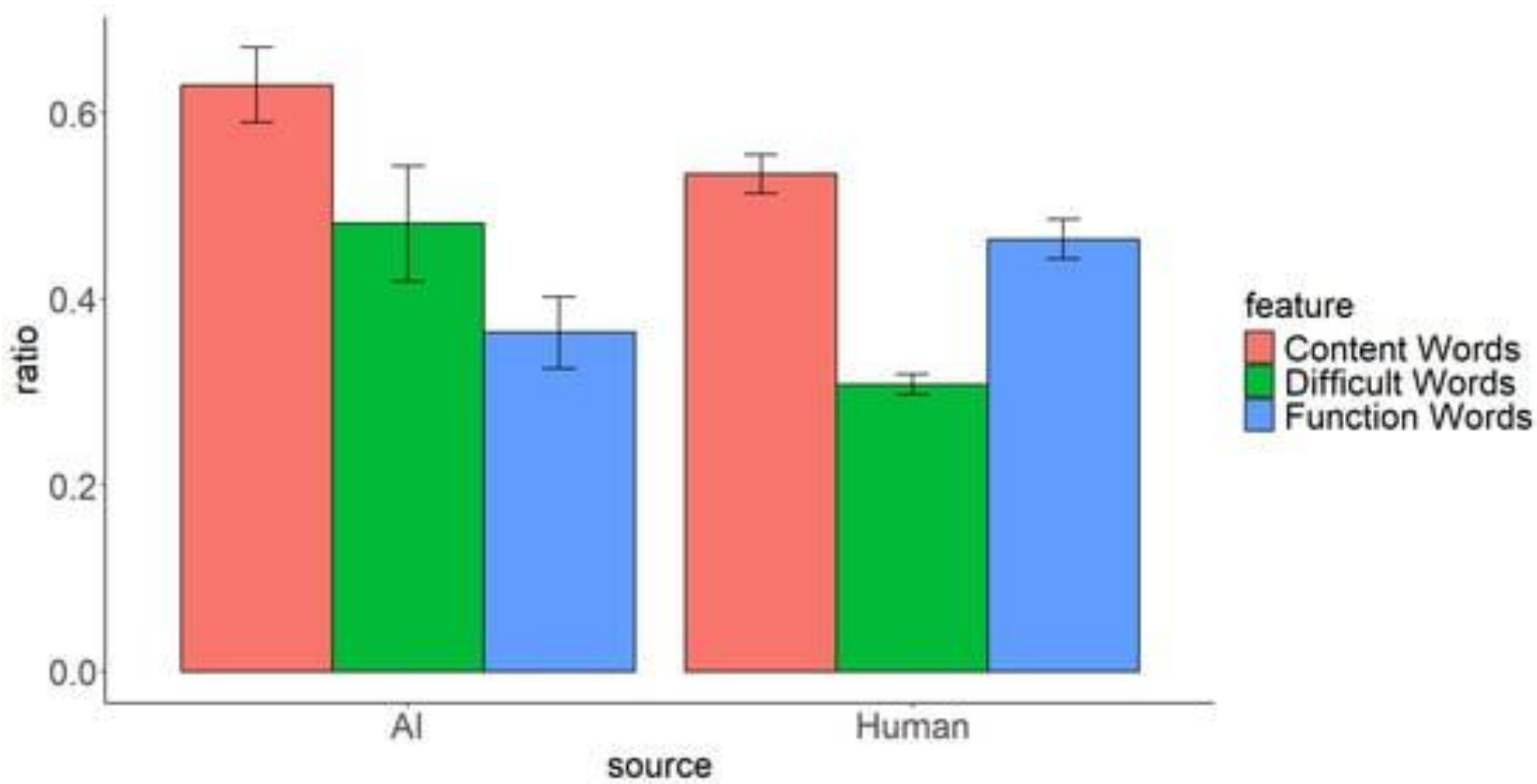

**Figure 5.** Ratios of lexical components in both the human-written and the AI-generated texts.

To identify the linguistic features that best distinguish AI-generated from human-authored texts, a LASSO model was employed. LASSO performs both variable selection and regularization by penalizing the absolute size of regression coefficients, effectively shrinking less informative predictors toward zero while retaining those that contribute most strongly to classification accuracy. In this analysis, positive coefficients indicate features occurring proportionally more often in AI texts, while negative coefficients indicate features more frequent in human texts. At the phonological level, primary stress carried the strongest negative weight, indicating a substantially higher ratio in human speech patterns. Conversely, alveolar and voiceless sounds were weighted positively, suggesting greater representation in AI output. Smaller negative weights for dental and neutral weights near zero for other features (e.g., approximant, fricative, nasal) indicate limited discriminative influence after regularization. In morphology, pronoun and auxiliary weights were strongly negative, highlighting heavier functional morphology in human texts, whereas noun was positive, indicating greater nominal density in AI writing. Other morphological categories were effectively neutral, suggesting a minimal role in distinguishing the two text types. At the syntactic level, the LASSO assigned a negative weight to prepositional modifiers, again suggesting higher frequency in human writing, while adjectival modifiers and conjuncts showed positive weights, reflecting their stronger presence in AI texts. Finally, within the lexical domain, function words were negatively weighted, consistent with human reliance on grammatical scaffolding, whereas difficult words were positively weighted, indicating slightly more complex vocabulary in AI output. The results of the LASSO model are presented in Table 3.

**Table 3.** Results of the LASSO model

| **level** | **feature** | ***p*-value** |
|---|---|---|
| phonology | primary stress | –269.2275 |
| | alveolar | 122.52697 |
| | voiceless | 53.26128 |
| | dental | –20.5736 |
| | approximant | 0 |
| | fricative | 0 |
| | lateral | 0 |
| | nasal | 0 |
| | plosive | 0 |
| | labiodental | 0 |
| | postalveolar | 0 |
| | voiced | 0 |
| | secondary stress | 0 |
| morphology | pronoun | –99.27591 |
| | auxiliary | –21.29675 |
| | noun | 21.17488 |
| | adjective | 0 |
| | adposition | 0 |
| | adverb | 0 |
| | coordinating conjunction | 0 |
| | particle | 0 |
| | subordinating conjunction | 0 |
| | verb | 0 |
| syntax | prepositional modifier | –26.1565252 |
| | adjectival modifier | 24.0081669 |
| | conjunct | 21.642647 |
| | nominal subject | 0.5338395 |
| | adverbial modifier | 0 |
| | determiner | 0 |
| | direct object | 0 |
| | object of preposition | 0 |
| | root | 0 |
| lexicon | function words | –8.95668793 |
| | difficult words | 0.01683913 |
| | content words | 0 |

## 4. Discussion

This study compared the occurrence of various phonological, morphological, syntactical, and lexical constituents in human-written versus AI-generated essay texts. The goal was to identify critical linguistic features that distinguish between the two types of texts. Linguistic features were elicited through an online platform application that has the capacity to extract linguistic information from written texts, among others. This is one of the few studies quantifying linguistic constituents produced by human and AI languages.

Our findings indicated that AI-generated text tended to favor the use of nasals, alveolars, and voiceless consonants more than human text; the latter two were major contributing factors according to the LASSO analysis. This preference was statistically significant with large effect sizes. Significant differences were observed in the usage of dentals and labiodentals, with AI text using these consonants less frequently. These trends warrant further investigation. The training data provided to AI models might influence their stylistic choices, potentially favoring sentences with a higher consonant density or stronger voiceless patterns. Interestingly, although primary stress did not differ significantly between AI-generated and human texts in the Wilcoxon rank-sum test, it emerged as the most influential phonological predictor in the LASSO model. This pattern suggests that primary stress does not independently separate the two groups but interacts with other phonological cues to form a characteristic profile of AI versus human prosodic structure. It is concluded that the internal algorithms governing AI text generation might prioritize specific phonological features during the construction of sentences. These results suggest that AI models may have inherent biases in consonant patterning, possibly due to the training data or the specific algorithms used. Such differences highlight an area where AI text generation could be fine-tuned for more human-like phonological characteristics. According to Suvarna et al. [31], the acquisition of phonological skills by LLMs is still in doubt since there is no access to speech data. The authors suggest that although LLMs perform well in tasks such as songwriting, poetry generation, and phonetic transcription, they lack deep phonological understanding. It should be noted that the phonological measures used in this study were inferred from orthographic representations rather than acoustic data. Consequently, our results rely on the accuracy of the grapheme-to-phoneme (G2P) conversion, which may introduce systematic biases, especially for words with irregular spellings or reduced forms. The phonological tendencies we report should therefore be interpreted as approximations of underlying sound patterns rather than direct evidence of phonetic realization. Future work integrating acoustic corpora or validated G2P mappings could refine these inferences and test the robustness of the observed contrasts.

The analysis of morphological features showed significant variations in the usage of adjectives, adpositions, auxiliaries, coordinating conjunctions, nouns, and pronouns. Notably, the AI text employed more coordinating conjunctions and nouns (the latter was a highly contributing feature), while the human text employed more adjectives, adpositions, auxiliaries, and pronouns (the latter two being highly contributing based on the LASSO analysis). These differences, with large effect sizes, may reflect the AI's tendency to produce more noun-heavy and conjunction-rich sentences, possibly making the text appear more formal or structured. The above findings are consistent with the results of Liao et al. [13], who reported greater usage of nouns and coordinating conjunctions by ChatGPT-generated compared to human-written medical texts. In addition, pronouns did not even appear in ChatGPT's top 20 for radiology, indicating relatively more pronouns in human writing. In general, our findings indicate that the human text, with its usage of adjectives, adpositions, auxiliaries, and pronouns, is syntactically and referentially rich, reflecting natural discourse features such as cohesion, perspective, and descriptive nuance. These findings suggest that while AI-generated texts can closely mimic human language, there are still distinct morphological patterns that differentiate them from human writing.

Moreover, the syntactic analysis uncovered significant differences in the use of adjectival modifiers, conjuncts, object prepositions, and prepositional modifiers. AI text was found to employ a higher number of tokens for the first two categories. The effect sizes were also large. The AI's preference for more conjuncts may contribute to a more segmented and explicit sentence structure, while human text's greater use of prepositional phrases, which ranked high in the model, could reflect a more relational and contextually embedded style. According to the lexical analysis, AI text tends to use more difficult words and content words, whereas human text is inclined to use function words, a feature that ranked high in the LASSO model. More advanced vocabulary of AI text compared to human text was also found by Alexander et al. [10]. This difference highlights the contrasting approaches in vocabulary selection, with AI possibly generating more sophisticated and varied vocabulary, while human authors may prioritize clarity and accessibility. Our finding that AI text shows higher lexical density and greater use of content words aligns with its tendency to emulate academic/technical registers, which are characterized by denser lexis and conventionalized phrasings (e.g., nominal style, recurrent bundles). Corpus-based register studies document that university/academic prose systematically prefers dense noun phrases, formulaic sequences, and informational packing compared with conversational styles, patterns our AI outputs appear to mirror, potentially making them feel more standardized and "formal" than human essays of similar length [16,17]. Therefore, ChatGPT achieves formality primarily through nominal and lexical density, not through increased passivization.

These patterns can also be interpreted through the lens of register variation. Following Biber and Conrad's multidimensional framework, our findings suggest that AI-generated texts align with features of informational/academic registers—higher lexical density, nominalization, and explicit cohesion devices—whereas human essays show more involved/interactive traits through greater use of pronouns and auxiliaries [32]. From a computational perspective, these regularities are consistent with token-probability bias and local-entropy/typicality dynamics in LLMs: standard decoding favors higher-probability continuations, yielding smoother entropy profiles and more regular POS distributions [33,34]. Visualization tools such as GLTR explicitly show a higher proportion of high-likelihood tokens in model outputs relative to human text [35]. Empirical comparisons further indicate that human writing regularly includes lower-probability tokens and greater local variability—differences that certain sampling strategies can narrow but not eliminate [36]. Conceptually, this reflects distinct production goals: human intentionality and audience design versus models' next-token optimization, which prioritizes distributional coherence over purpose-sensitive choice.

By examining linguistic patterns in the AI-generated language, LLMs can be trained in such a way as to improve their language development capabilities. This process involves scrutinizing various linguistic constituents within the generated text. By identifying and understanding these patterns, developers can adjust the training algorithms to better mimic natural language usage. This advancement has the potential to benefit various domains significantly. For example, in healthcare, chatbots offer valuable medical advice and guidance to individuals. The World Health Organization's technology program, for instance, has developed a chatbot to assist in combating COVID-19 [37]. This chatbot delivers information on virus protection, provides access to the latest news and facts, and helps users prevent the spread of the virus. Therefore, it is crucial for the language used by these chatbots to be as precise as possible.

Automatically and effortlessly eliciting linguistic features through an intuitive tool is crucial. Advances in AI have made it possible to seamlessly gather linguistic data from speakers. For example, Open Brain AI, which relies on AI and computer technology, provides a convenient tool for analyzing written texts. Through this online tool, we managed to extract measures of phonology, morphology, syntax, and lexicon of human and AI texts. This can be particularly useful to educators and clinicians [26]. Educators can aid students with speech, language, and communication challenges by using automated AI tools for assessment in these areas. Furthermore, clinicians can promptly screen and assess individuals with disorders, considering that early diagnosis can help prevent or slow the progression of these conditions. There are also economic benefits, as automation can reduce costs by requiring less effort and time to extract the data [27].

## 5. Conclusions

The results of this study have several implications for the development and refinement of AI text generation models. Overall, while ChatGPT-generated texts exhibit a high degree of linguistic competence, there are still discernible differences that set them apart from human writing based on the automated measures we gathered from Open Brain AI. The observed differences in phonology, morphology, syntax, and lexicon underscore the need for more refined training approaches that can produce more human-like text.

While feature-based detection approaches hold promise for explainable identification of AI-generated text, they are subject to practical and temporal limits. As models evolve and decoding strategies diversify, the linguistic signatures we identify may shift or attenuate, a phenomenon often referred to as a domain shift. Detection systems built on current surface-level cues risk rapid obsolescence and false positives when applied to newer model generations or different genres. Incorporating explainable AI methods—such as SHAP or LIME visualizations over classifier features—can mitigate this risk by making model decisions transparent and interpretable. However, these methods must be continuously recalibrated to reflect ongoing linguistic and algorithmic changes.

Building on these findings, future work should scale to larger, more diverse corpora, incorporate multilingual datasets beyond English, and compare genre/register variation (e.g., academic prose, journalism, narrative, and social media) to test the stability of the observed contrasts. Practically, the resulting profiles can inform automated authorship verification systems as human-in-the-loop decision supports; guide AI literacy training that teaches students to recognize register, density, and cohesion issues in model drafts; and enable explainable AI in education by making detection rationales transparent to instructors and learners. Ethically, we caution that feature-based detection has inherent limits: as models evolve and decoding strategies change, surface cues may attenuate or shift, risking brittleness, false positives, and misuse.

## Funding

This study is supported by the Phonetic Lab of the University of Nicosia.

**Institutional Review Board Statement**

This study is a non-biomedical, anonymous educational survey with no intervention, biological material, or health data; under Cyprus' Law 150(I)/2001 establishing the National Bioethics Committee and its Code of Practice/Operational Guidelines (Κ.Δ.Π. 175/2005)—which limit Review BioethicsCommittees to biomedical research involving human subjects or theirbiological substances (and clinical trials/medical devices)—nationalbioethics approval is not required. This study did not require the participants to complete any experiments but engaged them in simple anonymous writing tasks; a formal approval was not required.

**Informed Consent Statement**

Informed consent for participation was obtained from all subjects involved in the study.

**Data Availability Statement**

The data presented in this study are available on request from the corresponding author. The data are not publicly available due to restrictions related to copyright and licensing of the human-written essay samples.

**Conflicts of Interest**

The author declares no conflicts of interest.

**References**

1. Kasneci, E.; Seßler, K.; Küchemann, S.; Bannert, M.; Dementieva, D.; Fischer, F.; Kasneci, G. ChatGPT for Good? On Opportunities and Challenges of Large Language Models for Education. *Learn. Individ. Differ.* **2023**, *103*, 102274. [Google Scholar] [CrossRef]

2. Zhou, H.; Gu, B.; Zou, X.; Li, Y.; Chen, S.S.; Zhou, P.; Liu, F. A Survey of Large Language Models in Medicine: Progress, Application, and Challenge. *arXiv* **2023**, arXiv:2311.05112. [Google Scholar]

3. Adeshola, I.; Adepoju, A.P. The Opportunities and Challenges of ChatGPT in Education. *Interact. Learn. Environ.* **2023**, *32*, 1–14. [Google Scholar] [CrossRef]

4. Javaid, M.; Haleem, A.; Singh, R.P. ChatGPT for Healthcare Services: An Emerging Stage for an Innovative Perspective. *BenchCouncil Trans. Benchmarks Stand. Eval.* **2023**, *3*, 100105. [Google Scholar] [CrossRef]

5. Kohnke, L.; Moorhouse, B.L.; Zou, D. ChatGPT for Language Teaching and Learning. *RELC J.* **2023**, *54*, 537–550. [Google Scholar] [CrossRef]

6. Koc, E.; Hatipoglu, S.; Kivrak, O.; Celik, C.; Koc, K. Houston, We Have a Problem!: The Use of ChatGPT in Responding to Customer Complaints. *Technol. Soc.* **2023**, *74*, 102333. [Google Scholar] [CrossRef]

7. Ouyang, L.; Wu, J.; Jiang, X.; Almeida, D.; Wainwright, C.; Mishkin, P.; Zhang, C.; Agarwal, S.; Slama, K.; Ray, A.; et al. Training Language Models to Follow Instructions with Human Feedback. *Adv. Neural Inf. Process. Syst.* **2022**, *35*, 27730–27744. [Google Scholar]

8. Brown, T.; Mann, B.; Ryder, N.; Subbiah, M.; Kaplan, J.D.; Dhariwal, P.; Neelakantan, A.; Shyam, P.; Sastry, G.; Askell, A.; et al. Language Models are Few-Shot Learners. *Adv. Neural Inf. Process. Syst.* **2020**, *33*, 1877–1901. [Google Scholar]

9. Chukwuere, J.E. Today's Academic Research: The Role of ChatGPT Writing. *J. Inf. Syst. Inform.* **2024**, *6*, 30–46. [Google Scholar] [CrossRef]

10. Alexander, K.; Savvidou, C.; Alexander, C. Who Wrote This Essay? Detecting AI-Generated Writing in Second Language Education in Higher Education. *Teach. Engl. Technol.* **2023**, *23*, 25–43. [Google Scholar] [CrossRef]

11. Herbold, S.; Hautli-Janisz, A.; Heuer, U.; Kikteva, Z.; Trautsch, A. A Large-Scale Comparison of Human-Written versus ChatGPT-Generated Essays. *Sci. Rep.* **2023**, *13*, 18617. [Google Scholar]

12. Cai, Z.G.; Duan, X.; Haslett, D.A.; Wang, S.; Pickering, M.J. Do Large Language Models Resemble Humans in Language Use? *arXiv* **2023**, arXiv:2303.08014. [Google Scholar]

13. Liao, W.; Liu, Z.; Dai, H.; Xu, S.; Wu, Z.; Zhang, Y.; Li, X. Differentiating ChatGPT-Generated and Human-Written Medical Texts: Quantitative Study. *JMIR Med. Educ.* **2023**, *9*, e48904. [Google Scholar] [CrossRef]

14. Mitchell, E.; Lee, Y.; Khazatsky, A.; Manning, C.D.; Finn, C. DetectGPT: Zero-Shot Machine-Generated Text Detection Using Probability Curvature. In Proceedings of the International Conference on Machine Learning, Honolulu, HI, USA, 23–29 July 2023; pp. 24950–24962. [Google Scholar]

15. Bao, G.; Zhao, Y.; Teng, Z.; Yang, L.; Zhang, Y. Fast-DetectGPT: Efficient Zero-Shot Detection of Machine-Generated Text via Conditional Probability Curvature. *arXiv* **2023**, arXiv:2310.05130. [Google Scholar]

16. Biber, D.; Conrad, S.; Cortes, V. Lexical Bundles in University Teaching and Textbooks. *Appl. Linguist.* **2004**, *25*, 371–405. [Google Scholar] [CrossRef]

17. Biber, D.; Barbieri, F. Lexical Bundles in University Spoken and Written Registers. *Engl. Specif. Purp.* **2007**, *26*, 263–286. [Google Scholar] [CrossRef]

18. Emara, I.F. A Linguistic Comparison between ChatGPT-Generated and Nonnative Student-Generated Short Story Adaptations: A Stylometric Approach. *Smart Learn Environ.* **2025**, *12*, 36. [Google Scholar] [CrossRef]

19. Jaashan, H.M.; Bin-Hady, W.R.A. Stylometric Analysis of AI-Generated Texts: A Comparative Study of ChatGPT and DeepSeek. *Cogent Arts Humanit.* **2025**, *12*, 2553162. [Google Scholar] [CrossRef]

20. Wang, S.; Cristianini, N.; Hood, B.M. Stylometric Comparison between ChatGPT and Human Essays. In Proceedings of the 18th International AAAI Conference on Web and Social Media, Buffalo, NY, USA, 3–6 June 2024. [Google Scholar]

21. Muñoz-Ortiz, A.; Gómez-Rodríguez, C.; Vilares, D. Contrasting Linguistic Patterns in Human and LLM-Generated News Text. *Artif. Intell. Rev.* **2024**, *57*, 265. [Google Scholar] [CrossRef]

22. Sankar Sadasivan, V.; Kumar, A.; Balasubramanian, S.; Wang, W.; Feizi, S. Can AI-Generated Text Be Reliably Detected? *arXiv* **2023**, arXiv:2303.11156. [Google Scholar]

23. Georgiou, G.P.; Kaskampa, A. Differences in Voice Quality Measures among Monolingual and Bilingual Speakers. *Ampersand* **2024**, *12*, 100175. [Google Scholar] [CrossRef]

24. Georgiou, G.P.; Panteli, C.; Theodorou, E. Speech Rate of Typical Children and Children with Developmental Language Disorder in a Narrative Context. *Commun. Disord. Q.* **2025**, *46*, 212–221. [Google Scholar]

25. Themistocleous, C. Open Brain AI: An AI Research Platform. In Proceedings of the Huminfra Conference, Gothenburg, Sweden, 10–11 January 2024; Volume 200, pp. 1–9. [Google Scholar]

26. Themistocleous, C. Open Brain AI. Automatic Language Assessment. In Proceedings of the Fifth Workshop on Resources and Processing of Linguistic, Para-Linguistic and Extra-Linguistic Data from People with Various Forms of Cognitive/Psychiatric/Developmental Impairments @ LREC-COLING 2024, Torino, Italy, 25 May 2024; pp. 45–53. [Google Scholar]

27. Georgiou, G.P.; Theodorou, E. Detection of Developmental Language Disorder in Cypriot Greek Children Using a Neural Network Algorithm. *J. Technol. Behav. Sci.* **2024**. [Google Scholar] [CrossRef]

28. Stamatatos, E. A Survey of Modern Authorship Attribution Methods. *J. Am. Soc. Inf. Sci. Technol.* **2009**, *60*, 538–556. [Google Scholar] [CrossRef]

29. Kestemont, M. Function Words in Authorship Attribution: From Black Magic to Theory? In Proceedings of the 3rd Workshop on Computational Linguistics for Literature, Gothenburg, Sweden, 26–27 April 2014; pp. 59–66. [Google Scholar]

30. R Core Team. *R: A Language and Environment for Statistical Computing*; R Foundation for Statistical Computing: Vienna, Austria, 2024. [Google Scholar]

31. Suvarna, A.; Khandelwal, H.; Peng, N. PhonologyBench: Evaluating Phonological Skills of Large Language Models. *arXiv* **2024**, arXiv:2404.02456. [Google Scholar] [CrossRef]

32. Biber, D.; Conrad, S. *Register, Genre, and Style*; Cambridge University Press: Cambridge, UK, 2009. [Google Scholar]

33. Holtzman, A.; Buys, J.; Du, L.; Forbes, M.; Choi, Y. The Curious Case of Neural Text Degeneration. In Proceedings of the International Conference on Learning Representations (ICLR 2020), Addis Ababa, Ethiopia, 26–30 April 2020. [Google Scholar]

34. Meister, C.; Pimentel, T.; Wiher, G.; Cotterell, R. Locally Typical Sampling. *Trans. Assoc. Comput. Linguist.* **2023**, *11*, 102–121. [Google Scholar] [CrossRef]

35. Gehrmann, S.; Strobelt, H.; Rush, A.M. GLTR: Statistical Detection and Visualization of Generated Text. In Proceedings of the 57th Annual Meeting of the Association for Computational Linguistics: System Demonstrations, Florence, Italy, 28 July–2 August 2019; pp. 111–116. [Google Scholar]

36. Sasse, K.; Barham, S.; Kayi, E.S.; Staley, E.W. To Burst or Not to Burst: Generating and Quantifying Improbable Text. *arXiv* **2024**, arXiv:2401.15476. [Google Scholar] [CrossRef]

37. Walwema, J. The WHO Health Alert: Communicating a Global Pandemic with WhatsApp. *J. Bus. Tech. Commun.* **2021**, *35*, 35–40. [Google Scholar] [CrossRef]